\documentclass[runningheads]{llncs}
\usepackage[T1]{fontenc}
\usepackage{graphicx}
\usepackage{amssymb}
\usepackage{amsmath}
\usepackage{graphicx}
\usepackage{array}
\usepackage{paralist}
\usepackage{booktabs}
\usepackage{comment}
\usepackage{multirow}
\usepackage{hyperref}
\usepackage{colortbl}

\usepackage[tableposition=top,font=small,labelfont=bf]{caption}

\usepackage[table,dvipsnames]{xcolor}

\usepackage{todonotes}
\usepackage[autolanguage]{numprint}
\usepackage{capt-of}

\usepackage{arydshln}
\begin{document}

\title{Evaluating LLM Abilities to Understand Tabular Electronic Health Records: A Comprehensive Study of Patient Data Extraction and Retrieval}
\titlerunning{Evaluating LLM Abilities to Understand Tabular EHR}

\author{Jesús Lovón-Melgarejo$^{1}$ \and Martin Mouysset$^{1}$ \and Jo Oleiwan$^{1}$ \and Jose G. Moreno$^{1}$ \and Christine Damase-Michel$^{2}$ \and Lynda Tamine $^{1}$ }


\institute{${}^{1}$Université Paul Sabatier, IRIT, Toulouse, France\\
${}^{2}$Centre Hospitalier Universitaire de Toulouse, CERPOP INSERM UMR 1295 - SPHERE team, Faculté de Médecine Université de Toulouse, Toulouse, France\\
{\email{\{jesus.lovon, martin.mouysset, jo.oleiwan, jose.moreno,tamine\}@irit.fr}}\\
{\email{\{christine.damase-michel\}@univ-tlse3.fr}}
}


\authorrunning{Author}

\maketitle              
\begin{abstract}
Electronic Health Record (EHR) tables pose unique challenges among which is the presence of hidden contextual dependencies between medical features  with a high level of data dimensionality and sparsity. 
This study presents the first investigation into the abilities of LLMs to comprehend EHRs for patient data extraction and retrieval. We conduct extensive experiments using the MIMICSQL dataset to explore the impact of the prompt structure, instruction, context, and demonstration, of two backbone LLMs, Llama2 and Meditron, based on task performance. Through quantitative and qualitative analyses, our findings show that optimal feature selection and serialization methods can enhance task performance by up to 26.79\% compared to naive approaches. Similarly, in-context learning setups with relevant example selection improve data extraction performance by 5.95\%. Based on our study findings, we propose guidelines that we believe would help  the design of LLM-based models to support health search.  

\end{abstract}
\section{Introduction}
Large Language Models (LLMs)' applications on tabular data \cite{Sui2024,fang2024,Ye2023,Hegselmann2022TabLLMFC}  (e.g., question-answering \cite{Hall2018} and table search \cite{Trabelsi22}) 
are beneficial in several domains such as finance \cite{Deng23} and health \cite{STEINBERG2021}. However, recent studies \cite{fang2024,Sui2024} show that the gap between tabular data and natural language significantly hinders LLM's performance on downstream tasks due to several challenges among which  lack of standard adequate transformation from data to text, lack of grounding with prior knowledge, 
and lack of generalizability across data structures.\\
In particular, Electronic Health Record (EHR) tables include high-level of data heterogeneity, high dimensionality, and sparsity. Indeed, an EHR is a digital patient's medical history that contains vast amounts of heterogeneous tables (e.g., 26 tables in MIMIC-III \cite{Johnson2016}) that comprise a temporal and longitudinal structure of patient visits. Each visit includes both administrative and clinical data related to a significant number of features with different natures (e.g., diagnosis, procedure and medication codes, dosages). While standard health-related tasks such as information extraction and retrieval require a synergic understanding of structure and content from one patient-entity view, previous work studied, agnostically to domain application, the capabilities of LLMs to understand separately table structure (e.g., table splitting and parsing \cite{Sui2024}) and tasks (e.g., classification, question-answering \cite{Ye2023,Hegselmann2022TabLLMFC}) using tables as a set of elements. Thus, it is still unclear whether previous findings on LLMs' understanding of tabular data are transferable to EHRs and patient-related tasks. Furthermore, there are so far no clear findings about the extent to which LLM's prompt elements such as \textit{instruction}, \textit{context}, and \textit{demonstration} intertwine to jointly impact the performance of LLM's in health search tasks, namely extraction and retrieval. To sum up, there is a critical gap in the literature regarding standard best practices and guidelines  for prompting LLMs on EHR-related tasks. \\
In this paper, we aim to fill this gap. To achieve this goal, we conduct extensive experiments using the publicly available MIMICSQL\footnote{https://github.com/wangpinggl/TREQS} \cite{Wang20} dataset, based on the MIMIC III benchmark dataset widely adopted in the literature \cite{Johnson2016}. 
By exploring the effect of \textit{instruction}, \textit{context},  and \textit{in-context demonstration} on  task performance, our study consists of the following highlights: 1) investigating the LLMs' understanding of the relationship between EHR structure and content by evaluating the joint impact of EHR serialization and medical feature selection; 2) analyzing the effect of providing guided vs. non-guided instructions regarding the task outcome; and 3) measuring the impact of in-context demonstration quality  within an In Context Learning (ICL) setup on task performance. Our findings lead us to assess the following: 1) synergic comprehension of content and structure through EHR serialization and feature selection is significantly sensitive to prompt context  with improvements up to $26.79\%$. In particular, LLM' self-generated EHR table descriptions are more impactful on task performance;  2) the use of guided instructions has minimal impact on task performance; 3) ICL positively impacts LLMs' performance, particularly for the extraction task, with the best results obtained by selecting examples that better match the query input rather than the patient; and 4) LLMs have more difficulty retrieving relevant tabular data than extracting relevant tabular data from EHRs. 
We summarize our contributions as follows:
\begin{itemize}
    \item Through an extensive empirical study using standard medical datasets, we provide researchers and practitioners guidelines for suitably prompting LLMs on EHR-related tasks;
    \item We propose two new datasets MIMIC$_{ask}$ and MIMIC$_{search}$, based on the MIMICSQL dataset, to benchmark LLMs' on health data extraction and retrieval; 
    \item We provide for future work, code implementation for EHR data-to-text transformation techniques, instruction formatting, and patient example selection strategies\footnote{\url{https://github.com/jeslev/llm-patient-ehr}}, as well as corresponding baselines per task.
    
\end{itemize}

\section{Prompting LLMs on tabular data}
 Early work dealing with data-to-text generation focused on the design of suited structure-aware encoder-decoder architectures \cite{deng2022turl,puduppully-etal-2019,Liu2017,rebuffel2020hierarchical} and specific pretraining strategies  (e.g.,  TaPas \cite{deng2022turl}, TUTA \cite{Wang2021},  and UTP \cite{chen2023bridge}). Recent efforts leverage the strengths of decoder-only architectures through LLMs  \cite{fang2024}. The core underlying issue is the design of appropriately structured prompts composed of linear texts describing the input data and in-context demonstrations for helping LLMs understand the outputs \cite{Wei2022EmergentAO}. \\
To set up effective techniques that convert tabular data into linear texts fed to LLMs, previous work has relied on 1) hand-crafted templates using JSON, HTML, XML, and X-separate formats \cite{singha2023tabular,Sui2024,lovon2024}; 2) embedding-based serialization techniques that rely on table encoders (e.g., UniTabPT \cite{Sarkar2023TestingTL} and TableGPT \cite{gong-etal-2020-tablegpt}); 3) graph-based serialization techniques that convert a table into a tree represented as a tuple fed to the LLMs \cite{Zhao2023LargeLM}; and 4) LLM self-generated table description \cite{Sui2024}. 
Overall, research findings show that LLMs' performance is heavily sensitive to prompt formats \cite{Sui2024,singha2023tabular,Hegselmann2022TabLLMFC}, and that most LLMs struggle to handle high-dimensional tables due to the long context they induce. This leads to a significant challenge in balancing between effectiveness and efficiency in terms of memory and computational cost \cite{liu-etal-2023-jarvix,Sui2023TAP4LLMTP}. ICL \cite{Wei2022EmergentAO} has also been shown to be impactful on the performance of tasks involving tabular data regardless of downstream tasks \cite{Sui2024,Chen2022LargeLM,Hegselmann2022TabLLMFC}. It has been shown that performance is optimized with a limited number of examples \cite{Chen2022LargeLM}.  Zhao et al. \cite{Zhao2023LargeLM} and Ye al.  \cite{Ye2023} also demonstrated the benefits of applying chain of thought reasoning (COT) to enhance search performance on tables.


\section{Study Design}
\subsection{Tasks}
We focus on tasks leveraging a repository $R$ that contains raw tabular data related to $n$ EHRs represented using a reference set of demographic and clinical features $F=\{f_1, \dots f_{k} \}$. The EHR of patient $p_i$  (with $1 \leq i \leq n$)  can be formalized as a reference table $T_i$ structured using a subset of features $F^{p_i} \subseteq F$ where $F^{p_i}=\{f^{p_i}_1, \dots f^{p_i}_{k_i} \}$, with $k_i$ is the number of EHR features in $T_i$. 
We assume that feature names are natural language strings (e.g., ``age'', ``blood pressure''). In practice, each table $T_i$ is built upon a subset of tables in $R$. 
We formally define two pilot tasks that we address in our work.   
\begin{itemize}
    \item \textit{Extraction}: Given  the EHR of patient \textbf{p$_i$} represented with table $T_i$, the goal is to generate an answer \textbf{a} to a natural language query \textbf{q} based on information in $T_i$, e.g.,  \textbf{q:} \textit{``specify the primary disease and icd9 code of patient id 1875''}; \\
    \item \textit{Retrieval}: The goal is to provide a list of EHR tables $T_1,\dots, T_m$ from the repository $R$, that are relevant to a given natural language query \textbf{q},  e.g., \textbf{q:} \textit{``Which male patients had done the lab test renal epithelial cells?''}. 

\end{itemize}

\subsection{Prompt design}

\subsubsection{Prompt format.} We consider hard prompts as triplets following the generic format composed of the concatenation of elements in the form <\texttt{Instruction [Demonstration] Context>}, where the symbol \texttt{[...]} indicates that the occurrence of the element is optional.  The element order of the format follows recommendations from previous findings  \cite{fang2024,Sui2023TAP4LLMTP}, where: 
\vspace{-0.2 cm}
\begin{itemize}
\item \texttt{Instruction} refers to a short textual description of the task  $I_e$ or $I_r$
for respectively the \textit{extraction} and \textit{retrieval} 
task. 

\item  \texttt{Context} includes two elements: 1) the natural language description $C_i$ such as $C_i=\phi(T_i)$  where $T_i$ is the EHR tabular form of the input patient \textbf{p$_i$}, 
\textbf{$\phi$} is a table serialization function; and 2) query \textbf{q} involved in the task (§3.1).

\item \texttt{Demonstration} comprises examples appended to the prompt within the ICL setting. Formally, we build a database $E$ of labeled examples   $(q, p, a)$ and $(q, a)$  for respectively the \textit{extraction} and \textit{retrieval} task, where $a$ is the gold answer for query $q$, and $p$ is the target patient for the \textit{extraction} task.  
The core component of the demonstration selection strategy relies on a retrieval function $\mathcal{\sigma}$ which provides high-quality examples from database $E$ to be fed as demonstrations to the LLM. 
\end{itemize}

\subsubsection{Prompting strategies.} Our strategies are given by the multiple configurations of the prompt format defined as follows.
\paragraph{\textit{Instruction.}} 
We follow recommendations from previous work that emphasize the positive impact of guided instructions \cite{slack2023tablet}. Specifically, we explore using \textit{Non-Guided} instructions vs. \textit{Guided} instructions through a step-by-step description of how the model should analyze the input to provide the expected output. 
\vspace{-0.3 cm}
\paragraph{\textit{Context.}} To investigate to which extent LLMs can understand EHR structure we explore a set of SOTA table serialization functions $\phi$ on  EHR table structure.
\begin{itemize}
   \item \textit{Templates (txt)} \cite{Hegselmann2022TabLLMFC,lovon2024}:  rows are converted into sentences using a simple text template for each column;
    \item \textit{X-separate (xsep)} \cite{Sui2023TAP4LLMTP,singha2023tabular}: rows are line-separated, columns separated with a special character. Particularly, we use HTML tags; 
   \item \textit{Self-generate (sgen)} \cite{wang-etal-2023-self-instruct,Sui2024}: following previous work, 
   we prompt a LLM to self-generate from an input EHR table a textual description with relevant features in function of a given question. 
\end{itemize}

\begin{figure}[t]
    \centering
    \includegraphics[width=\linewidth]{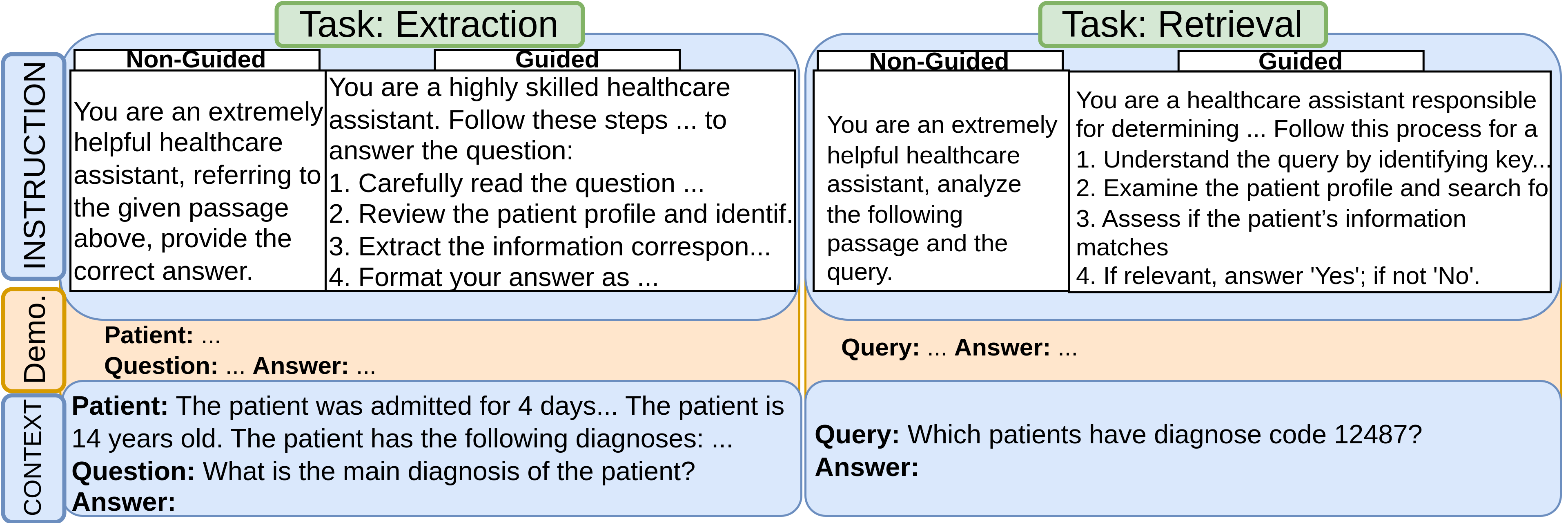}
    \caption{Illustration of the prompts used for the \textit{extraction} and \textit{retrieval}  tasks, including \textit{Guided} vs. \textit{Non-Guided} instructions, and patient with \textit{txt} (left) serializations.}
    \label{fig:prompts}
    \vspace{-2em}
\end{figure}

To synergically consider EHR table structure and content, we jointly explore a table structure serialization strategy $\phi$ with  table feature selection of a subset of medical features  $F^{p_i} \subseteq F$ as follows:
\begin{itemize}
   
   \item \textit{All (all):} all features and associated longitudinal values; 
   \item \textit{All aggregated (all$_{avg}$)}: all features and associated aggregated (averaged) longitudinal values;
    \item \textit{Random (rnd)}: random features and associated longitudinal values;  
   \item \textit{Random aggregated (rnd$_{avg}$)}\footnote{For both random approaches, we kept $60\%$ of the total features}: random features and associated aggregated (averaged) longitudinal values. 
\end{itemize}
\vspace{-0.3 cm}
\paragraph{Demonstration selection. } Previous work showed the impact of in-context example quality on downstream task performance \cite{rubin-etal-2022-learning,zhang-etal-2022-active}. We explore two main strategies of demonstration selection by varying the core  retrieval function $\mathcal{\sigma}$:  
\begin{itemize}
    \item \textit{Patient-based retrieval function} $\mathcal{\sigma}_p$ which retrieves high-quality examples from $E$ based on patient similarity. This strategy is applicable for the \textit{extraction} task which involves a patient input description \textbf{p} (§3.1).
    \item \textit{Query-based retrieval function} $\mathcal{\sigma}_q$ which retrieves high-quality examples from $E$ based on query similarity. This strategy is applicable for the \textit{extraction} and \textit{retrieval} tasks which both involve an input query \textbf{q} (§3.1).
\end{itemize}
Figure \ref{fig:prompts} summarizes the prompt formats used w.r.t each studied task (§3.1).
  
\subsection{Datasets}
Our study uses the MIMICSQL dataset \cite{Wang20}, based on the MIMIC III dataset, which contains de-identified EHRs from 48,520 ICU patients over a decade (2001-2012), structured into 26 tables. 
Each EHR record includes demographics and medical features (age, laboratory measurements, diagnoses, etc.). MIMIC III provides detailed time-series clinical features (e.g., blood pressure, heart rate) with variable time stamps (second, minutes), and formats, leading to a high level of heterogeneity and data sparsity. 
The MIMICSQL dataset \cite{Wang20} is a question-SQL pair dataset based on MIMIC III, to perform the Question-to-SQL generation task in the healthcare domain. It comprises 10,000 questions, expressed in natural language and SQL queries using 5 tables from the original database (Demographics, Diagnosis, Procedure, Prescriptions, and Laboratory tests).


We used the MIMICSQL dataset to perform the \textit{extraction} and \textit{retrieval} tasks and created the new MIMIC$_{ask}$ and MIMIC$_{search}$ datasets. We focused on the questions related to \textbf{single} and \textbf{multiple} patients to explore LLM abilities to comprehend EHRs, and omitted general questions that target database-level facts.  To build the ground truth, we evaluated the golden SQL queries provided in the original MIMICSQL dataset and generated corresponding question-query pairs by converting the SQL queries into their natural language form. We created upon the MIMICSQL dataset the MIMIC$_{ask}$ using \textbf{single} patient questions and MIMIC$_{search}$ datasets using \textbf{multiple} patient questions for the \textit{extraction} and \textit{retrieval} tasks respectively. For the MIMIC$_{ask}$ dataset, we cleaned the gold SQL answer by removing duplicates, serializing, and concatenating the features according to the format \textit{``column name: value''}. For the \textit{retrieval} task, we adapt the original question into queries by using rule-based query reformulations to target patients as outputs (e.g., transforming \textit{``Count the male patients that had done the lab test...''} into \textit{``Which male patients had done the lab test...''}).

Finally, we created training, validation, and test datasets for each task, ensuring, among other factors, that there was no overlap of patients between the training and test sets. For the \textit{extraction} task, we used the originally sampled question answer about 100 random patients from MIMICSQL. For the \textit{retrieval} task, we increased the corpus to 4000 random patients and created two versions, \textit{small} and \textit{full}, based on the number of test queries ($1101$ and $250$ respectively). We used the MIMIC$_{search}$ \textit{small} dataset for our study and shared both versions with the community for future work. 
Table \ref{tab:datasets} presents the statistics of the MIMIC$_{ask}$ and MIMIC$_{search}$ datasets.

\begin{table}[btp]
 
    \centering
       \resizebox{0.9\linewidth}{!}{
    \begin{tabular}{ccccccc}
    \toprule
    &\textbf{\# patients(n)} & \textbf{\# features(k)}&\textbf{\# k/n}&\textbf{\# train}&\textbf{\# dev}&\textbf{\# test}\\
    \midrule
     \textbf{MIMIC$_{ask}$ }& 100&5414&34&861&96&372\\
     \textbf{MIMIC$_{search}$}&4000&19970&557&2204&368&1101$^{full}$ | 250$^{small}$\\
     \textbf{MIMICSQL}&46520&32340&3912&8000&1000&1000\\
     \bottomrule
    \end{tabular}
    }
    \caption{Statistics of MIMIC$_{ask}$, MIMIC$_{search}$ and MIMICSQL datasets.}
    \label{tab:datasets}
 \end{table}

\subsection{Experimental Setup} 

\subsubsection{LLM Setups.}
For our evaluation we selected two main LLMs previously applied in patient-related  tasks \cite{lovon2024}: Llama2-7B\footnote{\url{https://huggingface.co/meta-llama/Llama-2-7b-chat-hf}} (Llama) and Meditron-7B (Meditron)\footnote{\url{https://huggingface.co/malhajar/meditron-7b-chat}}. The latter leverages further pre-training on medical PubMed scientific corpus. 
We particularly used a $4$-bit quantization configuration and a maximum context length of $4096$ tokens. Additionally, we explored the performance of the LLMs under a fine-tuning approach using the LoRa optimization strategy.

To perform the \textit{extraction} task, we follow previous work \cite{yu2023generate} and provide the question (query) \textbf{q} and the serialized text-based description $\phi(T_i)$ of EHR of patient \textbf{p$_i$} to the LLM to generate the answer \textbf{a}.  Regarding  the \textit{retrieval} task (§3.1), we follow previous work \cite{zhuang2024setwise} and employ a zero-shot LLM-based pointwise re-ranker.  A list of k-top relevant patients to the input query $q$ is retrieved in the first stage with an unsupervised retriever and then a  LLM is prompted to generate whether the input candidate patient is relevant to the query.   The re-ranking stage is repeated for each candidate relevant EHR $\phi(T_x)$, for $x=1 \dots k$ rounds. At each round, we provided the LLM with query $q$, concatenated with the text-based description $\phi(T_x)$ (§ Figure \ref{fig:prompts}). 

\subsubsection{Dense retrievers.}
In our work, the retrieval function $\mathcal{\sigma}$ (§3.2) relies on an off-the-shelf dense unsupervised retriever, agnostic to all the pilot tasks. Since we focus, on the joint impact of EHR structure and content on task performance, the retriever selects the top-K examples from $E$ based on several EHR representations similar to those used for serialization based on function $\phi$ (§3.2). Following recent work \cite{lin2023ra}, we encoded queries and text-based EHR using the Dragon+ encoder \cite{lin2023train} in a pre-processing step with the training set of the corresponding task. Then, we used FAISS,  implemented through the Pyserini framework \cite{Lin_etal_SIGIR2021_Pyserini}, to index and retrieve the most relevant examples based on similarity.  In addition, we design the random selection \textit{Random} as a lower-bound baseline to evaluate the effectiveness of the feature selection strategies.

\subsubsection{Baselines.}
We established various baselines for each of the tasks studied in our work: 1) for the \textit{extraction} task, we evaluated different generative models, including T5\footnote{\url{https://huggingface.co/google/flan-t5-large}} and BART\footnote{\url{https://huggingface.co/facebook/bart-base}} in a zero-shot setup (T5$_{0}$ and BART$_{0}$), as well as fine-tuned versions in the target task (T5$_{ft}$ and BART$_{ft}$); 2) for the \textit{retrieval} task, we considered both sparse rankers  such as BM25, and dense rankers: MonoBERT and MonoT5, which were finetuned for the target task. Finally, we reproduced the TREQS model \cite{Wang20}, executed the generated SQL queries based on our test sets, and post-processed them following the original work  \cite{Wang20}. In case of execution errors due to syntactically incorrect generated queries, we considered them as empty outputs. We then adapted the final SQL answer to match the expected output for our tasks (§3.1).

\subsubsection{Metrics.}
We used standard evaluation metrics appropriate for each task: 1) for the \textit{extraction} task, we used Rouge-1 (R-1) and BERT score (B$_{score}$ referring to the F-1 score); 2) for the \textit{retrieval} task, we used MAP and Recall@100 (R).

\section{Results}

\begin{table}[tpb]

    \centering
    \resizebox{0.7\textwidth}{!}{
    \begin{tabular}{cccccccccccc}
    \toprule
    && \multicolumn{5}{c}{\textbf{Llama}}&  \multicolumn{5}{c}{\textbf{Meditron}}\\
    \cmidrule(lr){3-7} \cmidrule(lr){8-12}
    $F^p$& $\phi$& \multicolumn{2}{c}{\textbf{Extraction}} 
      & \multicolumn{2}{c}{\textbf{Retrieval}} & \textbf{$\Delta$\%} & \multicolumn{2}{c}{\textbf{Extraction}} 
      & \multicolumn{2}{c}{\textbf{Retrieval}} & \textbf{$\Delta$\%}\\  \midrule
    &&B$_{score}$ & R-1 & MAP & R  & & B$_{score}$ & R-1 &   MAP & R  \\
    \cmidrule(lr){3-4} \cmidrule(lr){5-6} 
    \cmidrule(lr){8-9} \cmidrule(lr){10-11} 
     \cellcolor{blue!30} & txt &56.18 &22.84&9.30&32.19& \cellcolor{blue!30} $+26.79$& 56.21& 23.26&8.31&29.01 &  \cellcolor{blue!30} $+21.53$ \\ 
      \cellcolor{blue!30}  & xsep & 57.10&20.97 &\underline{9.80} &\underline{33.39} &  \cellcolor{blue!30} $+27.64$ & 52.10 & 14.94  &7.65&27.44 &  \cellcolor{blue!30} $+11.12$   \\
    \parbox[t]{2mm}{\multirow{-3}{*}{\cellcolor{blue!30} \rotatebox[origin=c]{90}{all}}}  & sgen & \underline{57.80} & 23.25& \textbf{9.84}&\textbf{33.62} &\cellcolor{blue!30}$+11.11$& 52.47 & 17.51&7.63 &24.67&\cellcolor{blue!30}$+4.59$\\ 
        \midrule
     \cellcolor{green!30} & txt &   56.86 & \underline{23.28} &8.25 &27.70  &  \cellcolor{green!30} $+22.44$ & \underline{57.26} & \textbf{25.89}  &\textbf{10.34}& \textbf{32.09}& \cellcolor{green!30} $+21.88$ \\
     \cellcolor{green!30}    & xsep &  57.30 & 21.84 & 7.98  &28.35& \cellcolor{green!30} $+19.06$ &   54.88 & 19.85  &8.14 &29.03 & \cellcolor{green!30} $+23.40$ \\
         \parbox[t]{2mm}{\multirow{-3}{*}{\cellcolor{green!30} \rotatebox[origin=c]{90}{all$_{avg}$}}}& sgen & \textbf{58.46} &\textbf{24.36}  &8.52 &32.00&\cellcolor{green!30}$+7.01  $&\textbf{57.79}&\underline{23.95}  & 8.05&29.10&\cellcolor{green!30}$+6.62$  \\
    \midrule
    \cellcolor{blue!30} & txt & 51.60 & 12.69  & 8.72 &28.83& -& 53.61 & 14.09 &    8.27&25.07  & -\\
      \cellcolor{blue!30} & xsep & 52.14 & 12.94 & 8.98 &25.71 &-&50.53& 11.60 &  7.32&25.39  & -\\
      \parbox[t]{2mm}{\multirow{-3}{*}{\cellcolor{blue!30} \rotatebox[origin=c]{90}{rnd}}} & sgen &55.89 &18.16&9.21 &31.67 &-&51.57&15.08&7.19&26.14&-\\
       \midrule
      \cellcolor{green!30}& txt &  51.76 & 12.91 & 8.34 & 27.73 & -&53.66 & 14.66 & \underline{10.30}  &\underline{30.91}  & -\\
      \cellcolor{green!30}  & xsep & 52.54 & 12.75 &  8.17 & 28.87 & - &50.83 & 12.69 &  7.69 & 23.53 & -\\
      \parbox[t]{2mm}{\multirow{-3}{*}{\cellcolor{green!30} \rotatebox[origin=c]{90}{rnd$_{avg}$}}}   & sgen  &56.58 &20.63&7.90&32.39&-&56.72&20.40&7.53&29.02&- \\
       \bottomrule
    \end{tabular}
     }
    \caption{Evaluation on the joint impact of EHR feature selection $F^p$, and EHR structure serialization $\phi$. $\Delta$ shows global improvement across tasks and metrics per setting (row) w.r.t their color corresponding baseline (\colorbox{blue!30}{all vs rnd}, \colorbox{green!30}{all$_{avg}$ vs rnd$_{avg}$}) . We report \textbf{best} and \underline{second best} values per metric (column).}
    \label{tab:features}

    
\end{table}

\subsection{Leveraging tabular EHR structure and content}

The results of our exploration of the joint impact of EHR feature selection ($F^p$) and structure serialization ($\phi$) on \textit{extraction} and \textit{retrieval} performances using the Llama and Meditron models are shown in Table \ref{tab:features}. 
At first glance, we can see that the optimal setting for Llama is using all features with \textit{sgen} serialization ($F^p=$all, $\phi=$sgen), achieving 3 out of 4 top scores, and Meditron using all aggregated features and \textit{txt} serialization ($F^p=$all$_{avg}$, $\phi=$txt).

Furthermore, we can notice that both models are sensitive to structure with a wide improvement variation range across feature selection strategies, with $\Delta$ from $7.01$ to $27.64$, and from $4.59$ to $23.40$ for Llama and Meditron, respectively. Also, the highest improvements are observed for Llama when all values are used ($F^p=$all and $F^p=$all$_{avg}$), while for Meditron only when these values are averaged ($F^p=$all$_{avg}$).
Focusing on the variability of this impact on performance across tasks, we can surprisingly see that the \textit{retrieval} task seems more difficult for both models. For instance Meditron reaches a minimal improvement of $0.39\%$ ($10.34$ vs $10.30$) for \textit{retrieval} vs. $1.75\%$ ($52.47$ vs $51.57$) for \textit{extraction} w.r.t their baselines. Similarly, Llama drops performance by $-2.33\%$ ($7.98$ vs $8.17$) for \textit{retrieval} while improving at least $3.32\%$ ($58.46$ vs $56.58$) for \textit{extraction}. This could be explained by the fact that \textit{retrieval} intrinsically requires more abilities to comprehend, at a coarse-grained level, the EHR structure and content to answer patient-profile-oriented queries, while \textit{extraction} questions explicitly focus on fine-grained specific patient features. Thus, LLMs struggle to match the query with relevant passages corresponding to 'cells' in the EHR table. 
However, both models achieve top performances on both tasks when using all features under different metrics.
Specifically, by cross-linking EHR structure, feature selection and task performance, we highlight from Table \ref{tab:features} that the Llama model with $F^p=$\{all,all$_{avg}$\} and \textit{sgen} method consistently outperforms  other settings, with one exception in the \textit{retrieval} task where \textit{xsep} also achieves high scores.
In contrast, Meditron shows a clear positive trend for $F^p=$all$_{avg}$ with \textit{txt} and \textit{sgen} methods. The preference of Meditron for \textit{txt} over \textit{sgen} suggests that textual medical knowledge captured by Meditron from the literature corpus endows it with better abilities to leverage EHR-related tasks with this same text format \textit{txt}. Interestingly, by comparing $F^p=$all and $F^p=$rnd$_{avg}$, we can see that Meditron exhibits a close gap in performance between these two settings, with better performance with $F^p=$rnd$_{avg}$ in the \textit{retrieval} task. This suggests that averaging longitudinal values positively impacts Meditron's performance, even using \textit{random} features.  

Overall this first exploration confirms findings from previous works about the sensitivity of LLM prompts on the performance of tabular tasks \cite{Sui2024,singha2023tabular,Hegselmann2022TabLLMFC}. It also reveals the following insights: 1) patient data \textit{retrieval} is a more difficult task than patient data \textit{extraction} for both LLMs; 2) Llama, a general domain LLM, lean to require all patient (\textit{all}) salient features (\textit{sgen}) while Meditron comprehends simple concatenation (\textit{txt}) of averaged patient feature values (\textit{all$_{avg}$}) to perform both \textit{extraction} and \textit{retrieval} tasks. \\
In the following sections, we chose the best settings from Table \ref{tab:features}: $(F^p$ =all, $\phi$ =sgen$)$ and $(F^p$ =all$_{avg}$, $\phi$ =sgen$)$ for Llama; and$(F^p$ =all$_{avg}$, $\phi$ =txt$)$ and $(F^p$ =all$_{avg}$, $\phi$ =sgen$)$ for Meditron.

\begin{table}[tpb]

    \centering
    \resizebox{\textwidth}{!}{
    \begin{tabular}{ccccccccccccccccc}
    \toprule
    & \multicolumn{8}{c}{\textbf{Llama}}&  \multicolumn{8}{c}{\textbf{Meditron}}\\
    \cmidrule(lr){2-9} \cmidrule(lr){10-17}
     ($F^p$,$\phi$) & \multicolumn{4}{c}{(all, sgen)} & \multicolumn{4}{c}{(all$_{avg}$,sgen)} & \multicolumn{4}{c}{(all$_{avg}$,txt)} & \multicolumn{4}{c}{(all$_{avg}$,sgen)} \\
    \cmidrule(lr){2-5} \cmidrule(lr){6-9} \cmidrule(lr){10-13} \cmidrule(lr){14-17}
    & \multicolumn{2}{c}{\textbf{Extraction}} 
      & \multicolumn{2}{c}{\textbf{Retrieval}}  & \multicolumn{2}{c}{\textbf{Extraction}} 
      & \multicolumn{2}{c}{\textbf{Retrieval}} & 
      \multicolumn{2}{c}{\textbf{Extraction}} 
      & \multicolumn{2}{c}{\textbf{Retrieval}}  & \multicolumn{2}{c}{\textbf{Extraction}} 
      & \multicolumn{2}{c}{\textbf{Retrieval}}\\  \midrule
    $I_e/I_r$&B$_{score}$ & R-1 & MAP & R  &  B$_{score}$ & R-1 &   MAP & R & B$_{score}$ & R-1 & MAP & R  &  B$_{score}$ & R-1 &   MAP & R  \\
    \cmidrule(lr){1-1} \cmidrule(lr){2-3} \cmidrule(lr){4-5} 
    \cmidrule(lr){6-7} \cmidrule(lr){8-9} \cmidrule(lr){10-11} \cmidrule(lr){12-13} \cmidrule(lr){14-15} \cmidrule(lr){16-17}
    Guided & \textbf{57.92} &\textbf{23.44} & 9.56& 33.42&\textbf{58.93}&\textbf{24.98}&\textbf{9.22} & 31.32&57.26&25.71&\textbf{12.07}&\textbf{32.54}&\textbf{57.82}&23.77  &7.98&28.95\\
Non-Guided &57.80 &23.25&\textbf{9.84}&\textbf{33.62}&58.46&24.36&8.52&\textbf{32.00}&57.26&\textbf{25.89}&10.34&32.09&57.79&\textbf{23.95}&\textbf{8.05}&\textbf{29.10}\\
       \bottomrule
    \end{tabular}
     }
    \vspace{-0.2cm} 
    \caption{Evaluation on the impact of guideline instructions on extraction and retrieval performance. \textbf{Bold} reports best score per metric (column).}
    \label{tab:instruction}

    \vspace{-1.5em}
    
\end{table}

\subsection{Guiding task completion}
Next, we evaluate the impact of the \textit{instruction} component, $I_e$ and $I_r$, by comparing two types: \textit{Guided} vs. \textit{Non-Guided}, based on the optimal settings identified in Section 4.1. Table \ref{tab:instruction} shows that the choice of instruction type has minimal impact on task performance, with no consistent improvement observed across all metrics.
Llama leverages \textit{Guided} instructions, achieving higher performance in $5$ out of $8$ metrics, compared to Meditron, which only shows improvements in $3$ out of $8$ metrics. This suggests that Llama benefits slightly more from explicit guidance, whereas Meditron’s performance may be hindered when adding detailed  \textit{Guided} instructions.
When comparing across tasks, we observe that \textit{Guided} instructions are particularly beneficial for the \textit{extraction} task ($I_e$), with $5$ out of $8$ metrics showing improvement, compared to $3$ out of $8$ for \textit{retrieval} ($I_r$). Notably, \textit{extraction} shows an average B$_{score }$ improvement of $0.27\%$, a trend inconsistent with other metrics. In conclusion, the choice of instruction type, $I_r$ and $I_e$, is heavily dependent on the model and setting with however limited impact on performance, unlike what has been found in previous work \cite{slack2023tablet}. \\
 For the next sections, we keep the following best-performing settings: $(F^p$ =all, $\phi =$sgen, \textit{Non-Guided} $)$ and $(F^p$ =all$_{avg}$, $\phi$ =sgen, \textit{Guided} $)$ for Llama; and $(F^p$ =all$_{avg}$, $\phi$ =txt, \textit{Guided}$)$ and $(F^p$ =all$_{avg}$, $\phi$ =sgen, \textit{Non-Guided}$)$ for Meditron.

\begin{table}[tpb]

    \centering
    \resizebox{\textwidth}{!}{
    \begin{tabular}{cccccccccccccccccc}
    \toprule
    && \multicolumn{8}{c}{\textbf{ICL w/ Llama (Lma$_{icl}$)}}&  \multicolumn{8}{c}{\textbf{ICL w/ Meditron (Med$_{icl}$)}}\\
    \cmidrule(lr){3-10} \cmidrule(lr){11-18}
     &($F^p$,$\phi$) & \multicolumn{4}{c}{(all, sgen)} & \multicolumn{4}{c}{(all$_{avg}$, sgen)} & \multicolumn{4}{c}{(all$_{avg}$, txt)} & \multicolumn{4}{c}{(all$_{avg}$, sgen)} \\
     &$I_e/I_r$ & \multicolumn{4}{c}{Non-Guided} & \multicolumn{4}{c}{Guided} & \multicolumn{4}{c}{Guided} & \multicolumn{4}{c}{Non-Guided} \\
     \cmidrule(lr){3-6} \cmidrule(lr){7-10} \cmidrule(lr){11-14} \cmidrule(lr){15-18}
    $\sigma$&\#ex & \multicolumn{2}{c}{\textbf{Extraction}} 
      & \multicolumn{2}{c}{\textbf{Retrieval}}  & \multicolumn{2}{c}{\textbf{Extraction}} 
      & \multicolumn{2}{c}{\textbf{Retrieval}} & 
      \multicolumn{2}{c}{\textbf{Extraction}} 
      & \multicolumn{2}{c}{\textbf{Retrieval}}  & \multicolumn{2}{c}{\textbf{Extraction}} 
      & \multicolumn{2}{c}{\textbf{Retrieval}}\\  \midrule
    &&B$_{score}$ & R-1 & MAP & R  &  B$_{score}$ & R-1 &   MAP & R & B$_{score}$ & R-1 & MAP & R  &  B$_{score}$ & R-1 &   MAP & R  \\
    \cmidrule(lr){3-4} \cmidrule(lr){5-6} 
    \cmidrule(lr){7-8} \cmidrule(lr){9-10} \cmidrule(lr){11-12} \cmidrule(lr){13-14} \cmidrule(lr){15-16} \cmidrule(lr){17-18}
- & 0 & 57.80&23.25&\textbf{9.84}&\textbf{33.62}&58.93&24.98&\underline{9.22}&31.32&57.26&25.71&\textbf{12.07}&32.54&\textbf{57.79}&\textbf{23.95}&\textbf{8.05}&\textbf{29.10}\\ \midrule
 &  1 & 59.47&	26.88&	N/A&		N/A&		60.75&	30.61&	N/A&		N/A&		57.08&	24.28&	N/A&		N/A&		\underline{56.61}&	\underline{21.84}&	N/A&N/A \\
 &2 &  59.97&	27.16&	N/A&		N/A&		60.51&	29.68&	N/A&		N/A&		54.60&	21.36&	N/A&		N/A&		51.10&	15.40&	N/A&		N/A \\
    \parbox[t]{2mm}{\multirow{-3}{*}{\rotatebox[origin=c]{90}{\scriptsize Patient$\sigma_p$}}}  
& 3 &60.75&	28.75&	N/A&		N/A&		60.75&	29.53&	N/A&		N/A&		54.92&	22.61&	N/A&		N/A&		50.72&	14.51&	N/A&		N/A \\ \midrule
    \parbox[t]{2mm}{\multirow{3}{*}{\rotatebox[origin=c]{90}{\scriptsize Query $\sigma_q$}}}
 &  1 &  60.82&	28.90&	8.06&	29.50&	60.36&	30.61&	7.84&	29.82&	\textbf{60.69}&	\textbf{33.18}&	10.15&	\underline{34.64}&	52.58&	18.49&	7.08&	24.33 \\
&2 &  \underline{61.42}&	\underline{30.04}&	8.07&	29.43&	\underline{61.90}&	\underline{31.87}&	8.81&	30.80&	\underline{59.16}&	\underline{28.09}&	7.57&	20.97&	51.28&	15.81&	6.88&	23.48 \\
& 3 &  \textbf{61.83}&	\textbf{31.48}&	8.28&	\underline{31.95}&	\textbf{62.44}&	\textbf{32.39}&	8.85&	\underline{34.17}&	54.77&	22.88&	7.29&	23.34&	51.05&	15.95&	7.08&	26.68 \\
        \midrule
        \parbox[t]{2mm}{\multirow{3}{*}{\rotatebox[origin=c]{90}{\scriptsize Random}}}
 &  1 &  58.66&	25.33&	7.87&	29.63&	59.89&	28.55&	7.98&	29.60&	57.80&	26.75&	\underline{10.92}&	\textbf{35.64}&	53.69&	19.26&	\underline{7.22}&	24.85 \\
        &2 &59.90&	26.60&	7.87&	30.20&	59.80&	27.76&	8.67&	31.15&	56.86&	24.74&	7.88&	21.33&	51.91&	15.96&	6.80&	23.49 \\
& 3 &  59.75&	26.69&	\underline{8.48}&	31.15&	60.91&	28.90&	\textbf{9.42}&	\textbf{35.54}&	52.40&	17.47&	7.77&	24.92&	50.86&	15.06&	7.13&	\underline{27.54} \\
       \bottomrule
    \end{tabular}
     }
 \caption{Evaluation on demonstration selection strategies and number of examples within an ICL setting. We report \textbf{best} and \underline{second best} values per metric (column).}
    \label{tab:icl}

    
\end{table}
\subsection{Selecting demonstrations}

Now, we evaluate the impact of demonstration quality in an ICL setup on task performance. To this end, we varied the demonstration retrieval function $\mathcal{\sigma}$ (§3.2) and the number of demonstrations fed to the LLM, $k=\{1,2,3\}$. To fit the LLM’s context length constraints, we truncate patient serializations as necessary. 
We compare the ICL performance w.r.t scores obtained in Section $4.2$ with no demonstrations ($k=0$).
Our results are reported in Table \ref{tab:icl}.

The first surprising observation from these results is that the Meditron model fails to leverage the ICL setup across most settings. The best scores are obtained in 5 out of 8 metrics with $k=0$ demonstrations, leading to dropping performance up to $39.60\%$ in MAP and $12.24\%$ in B$_{score}$ with ICL. This behavior suggests that demonstrations do not bring to Meditron additional relevant external knowledge beyond the domain-specific internal knowledge acquired during fine-tuning.
Analyzing ICL's impact at the task level, we observe that the \textit{extraction} task benefits more from the ICL setup. On the contrary, the \textit{retrieval} task   consistently underperforms compared to zero-shot, with a performance drop of $-9.5\%$ between the best ICL setting and $k=0$, except for one Llama setting $(F^p$ =all$_{avg}$, $\phi$ =sgen, Guided), which shows an improvement of $2.17\%$ MAP. 
These results suggest that LLMs struggle to reason over patient EHRs in \textit{retrieval} tasks, even with demonstrations, whereas their performance on \textit{extraction} tasks consistently improves, with a $+5.95\%$ B$_{score}$ increase using query-based demonstrations. This aligns with our previous findings (§4.1) about LLMs' challenges in comprehending EHR structure and content for \textit{retrieval} tasks.

Regarding the impact of the demonstration, selection functions $\mathcal{\sigma}$, we can observe that query-based demonstrations ($\mathcal{\sigma}_q$) obtain the highest scores in 9 out of 16 metrics, while patient-based demonstrations ($\mathcal{\sigma}_p$) obtain lower performance than random demonstrations. We further analyze the optimal number of demonstrations, focusing only on query-based demonstrations ($\mathcal{\sigma}_q$). We can observe that Llama benefits mainly from $k=3$ demonstrations, while Meditron with $k=1$. In general, we observe improvements up to $10.81\%$ in B$_{score}$ when finding the optimal number of demonstrations for the \textit{extraction} task.

\begin{figure}[tbp]
    \centering
    \includegraphics[width=\textwidth]{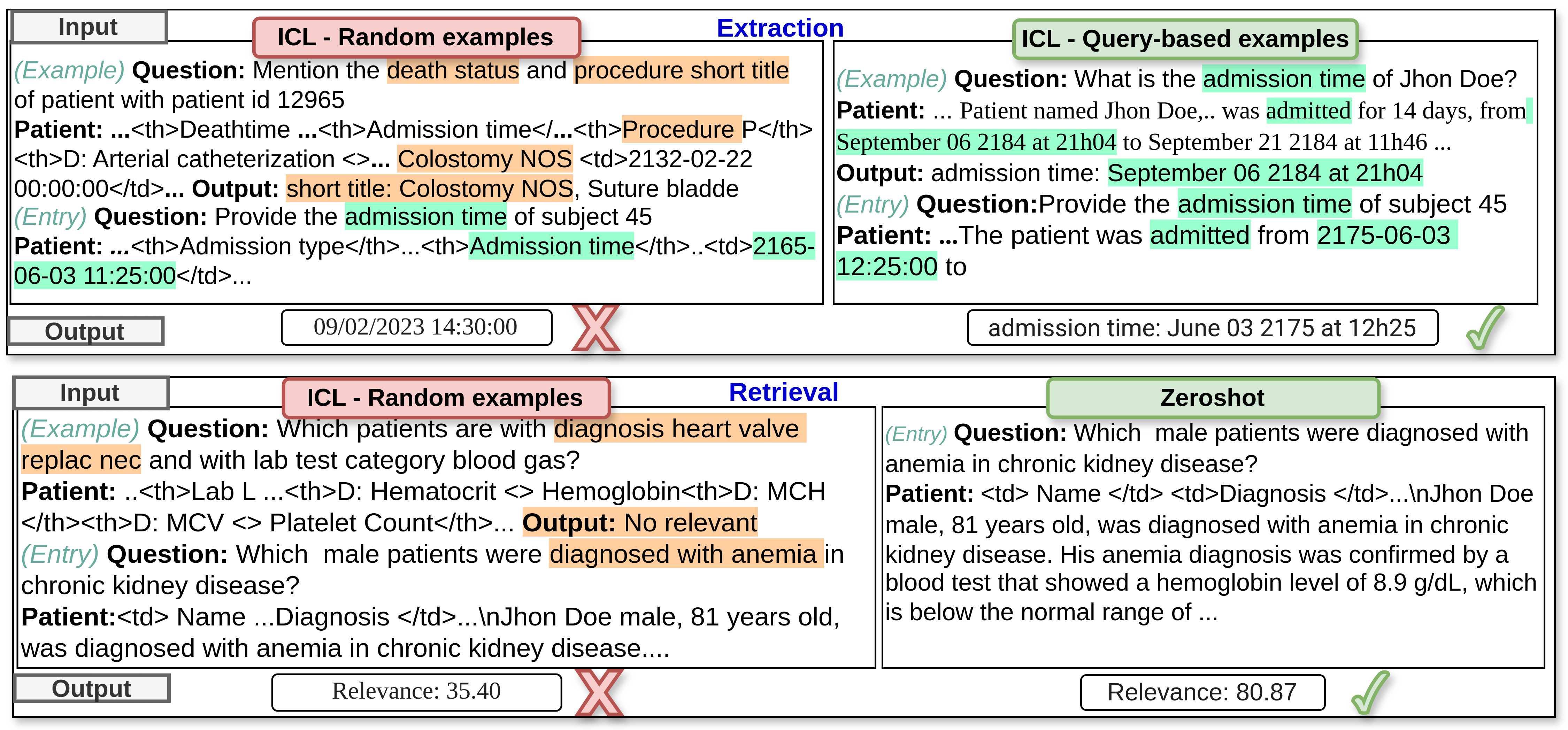}
    \vspace{-1 em}
    \caption{(top) Example of random and query-based demonstrations in an ICL setup for \textit{extraction}.  (bottom) Example of ICL and zeroshot setup for \textit{retrieval}. Highlighted the features (and values) referenced in demonstration and input.}
    \label{fig:examples}
\end{figure}

To better showcase this non-intuitive finding about the variability in the impact of ICL across tasks, we analyze selected examples of queries for each of the \textit{extraction} and \textit{retrieval} tasks based on the trends observed in Table \ref{tab:icl}. Figure \ref{fig:examples} shows a pair of such examples.
 We can observe that for the \textit{extraction} task, \textit{query-based} demonstrations share relevant features (highlighted) between the question and the patient's EHR profile (\textit{``admission time'', ``was admitted''}), guiding the LLM to find relevant information, particularly challenging for numerical information. In contrast, \textit{random} demonstrations lean to focus on irrelevant features, leading to nonfactual information generation. For the \textit{retrieval} task, we can observe that adding more demonstrations introduces more heterogeneous information into prompts (\textit{``diagnosis heart valve'', ``diagnosed with anemia''}), and conditioning the binary answer to the example output (\textit{``Output: No relevant''}), which appears to hinder the LLM's performance, whereas single clear EHR profiles obtain better relevance scores. 
 Overall our results assess that ICL significantly improves performance on \textit{extraction} tasks using query-based demonstration selection, but \textit{retrieval} benefits more from zero-shot setups. 

\begin{table}[tbp]
    \centering
    \resizebox{0.7\textwidth}{!}{
    \begin{tabular}{cccccccccc}
    \toprule
        \multicolumn{9}{c}{\textbf{Extraction}} \\ \midrule
          Models & T5$_0$       & BART$_0$      & T5$_{ft}$        & BART$_{fr}$    & TREQS        & Lma$^*$ & Med$^*$ & Lma$_{ft}^*$ & Med$_{ft}^*$\\ \midrule
\textbf{B$_{score}$} & 46.07  & 48.50 & 53.41 & 83.94  & 23.68&  62.44 &60.69&   \textbf{84.79}&  56.04     \\ 
\textbf{R-1} &  4.92 & 2.19  &  28.07  &67.18   & 13.21&       32.39&33.18&   \textbf{74.47}  &11.90   \\
\midrule
      \multicolumn{9}{c}{\textbf{Retrieval}} \\ \midrule
   Models& \multicolumn{2}{c}{BM25}       & MonoB  & MonoT5          &TREQS&Lma$^*$ & Med$^*$ & Lma$_{ft}^*$ & Med$_{ft}^*$\\ \midrule

\textbf{MAP}   &  \multicolumn{2}{c}{35.26}  & 10.19   & 38.49      &   43.99     & 9.84&  12.07&     11.33 &\textbf{44.34}  \\ 
\textbf{R}   & \multicolumn{2}{c}{49.37} &35.43  & 53.01  & 52.16 &      33.62 &32.54  &47.95& \textbf{53.38}\\ 

       \bottomrule
    \end{tabular}
    }
    \caption{Results for patient-related tasks using different SOTA models and best settings found for LLMs. All baselines were evaluated using ($F^p=$all$_{avg}$, $\phi=$txt, Non-Guided). We report \textbf{best} scores per metric (row). }
    \label{tab:sota}
\end{table}

\section{Comparative evaluation and guidelines}
To better support our final guidelines, we compare in Table \ref{tab:sota}, the optimal settings previously obtained, denoted  Lma$^*$ and Med$^*$ for Llama and Meditron, with: 1) different state-of-the-art models per task and 2) their fine-tuned counterparts, Lma$^*_{ft}$ and Med$^*_{ft}$, using the new MIMIC$_{ask}$ and MIMIC$_{search}$ datasets.  


We observe that for both tasks, fine-tuned LLMs achieve the highest scores, demonstrating their adaptability to leverage EHR tabular data when using optimal feature selection and serialization techniques. Specifically, for the \textit{extraction} task, we observe that explored LLMs perform better than all baselines, with the best score for Lma$^*$ with B$_{score}=62.44$, except BART$_{ft}$ with B$_{score}=83.94$. Interestingly, Med$_{ft}^*$ shows a performance decrease compared to the zero-shot counterpart Med$^*$, highlighting the challenge of fine-tuning LLMs for complex tasks \cite{yang2024unveiling}. Additionally, we can see that the TREQS model achieves low B$_{score}$($23.68$), mainly due to frequent errors in retrieving specific columns from the EHRs. For the \textit{retrieval} task, we note that studied LLMs only outperform MonoBERT, though BM25 and MonoT5 surpass them, emphasizing the difficulty of \textit{retrieval} tasks for LLMs as shown in the literature \cite{hou2024large}. Notably, only Med$_{ft}^*$ exhibits a better task comprehension, outperforming other models. Interestingly, the TREQS performance trend is reversed compared to the \textit{extraction} task, by achieving the second-best performance. This suggests model limitations to handle the sparsity of features to retrieve fine-grained clinical features in complex tabular data while being optimal to retrieve general coarse-grained conditions. In summary, our exploration provides the following guidelines about the use of LLMs for tabular EHR-related \textit{extraction} and \textit{retrieval} tasks: 

\vspace{-0.1 cm}
\begin{enumerate}
\item Context is improved when using all available EHR features, leading to better task performance. If longitudinal values are present in the EHRs, the best performance is reached by feature value aggregation;
\item The best EHR serialization method is based on the LLM self-generated EHR tabular descriptions, particularly for zero-shot LLMs. Medical pre-trained LLMs  can handle naive template-based serialization;
\item In an ICL setting, demonstration selection based on queries is more effective for extraction as the number of examples increases. Unlikely, the retrieval task better leverages zero-shot setups;
\item Finetuned LLMs with basic data-to-text EHR serialization methods achieve the best performance across tasks against fine-tuned pre-trained models.
\end{enumerate}

\section{Conclusion}
In this paper, we explored prompt techniques of LLMs on tabular EHR extraction and retrieval. Our results showed a trend toward retrieval being more challenging than extraction.  Our study has also shown that LLMs' performance on both tasks is particularly impacted by feature selection, serialization methods, and the quality of in-context demonstrations with significant levels of variations across tasks and backbone LLMs. Overall, our findings provide actionable insights for optimizing LLM performance on tasks involving tabular EHR data.\\
Our work has some limitations. We only consider two LLMs  based on the open Llama model. Thus, the generalization of these guidelines to other model architectures with different sizes and training data remains unexplored. Moreover, for computational limitations, we restricted the exploration to the optimal settings identified, omitting possible causality relationships that other settings would have revealed. 
In future work, we will extrapolate our study to predictive EHR-related tasks and investigate to which extent extraction, retrieval, and prediction tasks can be used as control tasks for assessing data privacy protection in LLMs.

\subsubsection{\ackname} This work has been supported by the In-Utero project funded by HDH (France) and FRQS (Canada).
This work was also granted access to the HPC resources of IDRIS under the allocation 2024-AD011015371 made by GENCI.

\newpage
%
%
%
%
\newpage
\bibliographystyle{splncs04.bst}
\bibliography{ecir2025}
%


\end{document}